\documentclass[3p]{elsarticle}

\usepackage{hyperref}

\journal{}

\biboptions{authoryear}

\usepackage{multirow}
\usepackage{graphicx}
\usepackage{subfig}
\usepackage{amsmath}
\usepackage{amssymb}
\usepackage{hyperref}
\usepackage{booktabs,tabularx}
\newcolumntype{x}{>{\centering\arraybackslash}X}
\usepackage{float}
\usepackage[]{algorithm2e}

\usepackage{tikz}
\usetikzlibrary{arrows,decorations.pathmorphing,backgrounds,positioning,fit,petri}
\usetikzlibrary{bayesnet}
\usetikzlibrary{mindmap,trees}

\newcommand{\wrt}{\mbox{w.r.t.} }

\newcommand{\C}{\mathcal{C}}
\newcommand{\N}{\mathcal{N}}
\newcommand{\E}{\mathbb{E}}
\newcommand{\bs}{\boldsymbol}
\newcommand{\bx}{\textbf{x}}
\newcommand{\bX}{\textbf{X}}

\newcommand{\by}{\textbf{y}}

\newcommand{\bz}{\textbf{z}}

\newcommand{\tr}{{\mbox{\scriptsize T}}}

\begin{document}

\begin{frontmatter}

\title{Bayesian Automatic Relevance Determination for Utility Function Specification in Discrete Choice Models}


\author[mymainaddress]{Filipe Rodrigues\corref{mycorrespondingauthor}}
\ead[url]{http://fprodrigues.com}

\cortext[mycorrespondingauthor]{Corresponding author}
\ead{rodr@dtu.dk}

\author[mysecondaryaddress]{Nicola Ortelli}
\ead{nicola.ortelli@epfl.ch}

\author[mysecondaryaddress]{Michel Bierlaire}
\ead{michel.bierlaire@epfl.ch}

\author[mymainaddress]{Francisco C. Pereira}
\ead{camara@dtu.dk}

\address[mymainaddress]{Technical University of Denmark (DTU), Bygning 116B, 2800 Kgs. Lyngby, Denmark}
\address[mysecondaryaddress]{\'{E}cole Polytechnique F\'{e}d\'{e}rale de Lausanne (EPFL), }

\begin{abstract}
Specifying utility functions is a key step towards applying the discrete choice framework for understanding the behaviour processes that govern user choices. However, identifying the utility function specifications that best model and explain the observed choices can be a very challenging and time-consuming task. This paper seeks to help modellers by leveraging the Bayesian framework and the concept of automatic relevance determination (ARD), in order to automatically determine an optimal utility function specification from an exponentially large set of possible specifications in a purely data-driven manner. Based on recent advances in approximate Bayesian inference, a doubly stochastic variational inference is developed, which allows the proposed DCM-ARD model to scale to very large and high-dimensional datasets. Using semi-artificial choice data, the proposed approach is shown to very accurately recover the true utility function specifications that govern the observed choices. Moreover, when applied to real choice data, DCM-ARD is shown to be able discover high quality specifications that can outperform previous ones from the literature according to multiple criteria, thereby demonstrating its practical applicability. 
\end{abstract}


\end{frontmatter}


\section{Introduction}
\label{sec:introduction}

Discrete choice models (DCM) provide a powerful framework for understanding user behaviour. By modelling user choices as functions of the alternative-specific characteristics and user attributes, DCMs allow researchers to predict users' future choices given a set of discrete alternatives and understand the behaviour process that governs their choices. Hence, it is without surprise that DCMs have become a widely adopted framework in various domains ranging from psychology to economics, thus making them one of the main work-horses for understanding user travel behaviour, consumer behaviour, and many other kinds of user choices. 

In practice, a fundamental part of applying the DCM framework consists in specifying the utility function for each alternative in the choice set, which are generally assumed to be known a priori. For the sake of interpretability, these utility functions are typically assumed to be linear functions of a set of explanatory variables. Although limiting at first sight, this linear framework can be made rather powerful by exploring variable transformations (e.g. log-transformations, Box-Cox transformations), one-hot encodings, piecewise linear representations, discretizations, interactions between variables, etc. However, all these modelling choices quickly raise the number of possible utility function specifications beyond manageable values for the modeller. On the other hand, given the central role of the utility functions in DCMs, it is essential to determine good specifications, at the risk of obtaining misspecified models and biased parameter estimates \citep{torres2011wrong}. As a consequence, a modeller often spends large portions of time seeking the ``best'' specification according to different criteria (e.g. convergence, log-likelihood, p-values), typically through a combination of trial-and-error and domain knowledge (\mbox{e.g.} economic theories). 

In this paper, we propose leveraging the Bayesian framework in order to automatically determine an optimal utility function specification from an exponentially large set of possible specifications in a purely data-driven manner. Although the proposed approach is not meant to be a complete replacement for expert intuition and domain knowledge, it is shown to provide key insights about the data that can help the modeller determine the utility function specification that best represents the observed choice data, which can ultimately lead to new understandings about the way people make choices in certain contexts. 

Based on the principle of Automatic Relevance Determination (ARD), as developed by \cite{tipping2001sparse} in the context of the Relevance Vector Machine and as widely used in the Gaussian Processes literature \citep{rasmussen2003gaussian}, we propose the use of a hierarchical prior on the preference parameters of each utility function in order to automatically determine their relevance for explaining the observed choice data. The key idea consists in jointly estimating the posterior distribution over the preference parameters, as well as the optimal values for the variances of the Gaussian priors over each possible explanatory variable to be included in each utility function specification. In order to ensure consistency among the selected variables, \mbox{i.e.} that either all or none of the dimensions corresponding to the representation of a given explanatory variable are selected, we propose tying the variance parameters of the Gaussian priors over the parameters that correspond to the same representation of a given choice attribute. Given the estimated optimal values for the variances of the Gaussian priors for a very large set of possible variable representations, a modeller can easily determine the most relevant attributes and corresponding representations for explaining a dataset of observed choices by simply selecting the variables for which the estimated prior variances are non-zero. 

Since exact Bayesian inference in the proposed DCM-ARD model is intractable, we propose the use of the variational inference framework. Namely, we develop an efficient approximate inference algorithm using doubly stochastic variational inference \citep{titsias2014doubly}. By combining the theory of variational inference with the theory of stochastic optimization, the proposed inference algorithm is able to approximate the true posterior distribution over the preference parameters with a tractable distribution and jointly estimate the optimal Gaussian prior hyper-parameters, while being able to scale to very large datasets with a very high number of dimensions. Although we focus on Multinomial Logit (MNL) models, the proposed approach can be extended to more complex models such as Mixed and Latent Class Logit models. 

The validity of the proposed automatic utility function specification framework is empirically demonstrated using both semi-artificial and real choice data. We begin by empirically demonstrating the ability of the proposed approach to discover the correct utility function specifications through an extensive series of experiments on simulated choice data based on the Swissmetro dataset \citep{bierlaire2001acceptance}. In particular, we manually specify a series of ``artificial'' (but realistic) utility function specifications of increasing complexity and, based on the Swissmetro dataset, we sample new artificial choices according to the manually-specified utility functions. Our empirical results show that the proposed DCM-ARD model is able to very accurately recover the ``true'' specifications that were used to generate the artificial choices, even in settings where the number of variables representations and transformations considered for each utility function is in the order of the thousands. 
Lastly, our empirical results on the real choices from the Swissmetro dataset demonstrate the potential of the proposed framework for discovering novel utility function specifications that can potentially outperform previous ones from the state of the art in terms of explanatory power and generalization to unobserved data. 

In summary, the main contributions of this paper are the following:
\begin{itemize}
\item We adapt the theory of ARD to the domain of DCMs, making the necessary modifications that are required from a choice modelling perspective (\mbox{e.g.} multiple utility functions with alternative-specific attributes, variable number of dimensions and tied parameters in the hierarchical priors);
\item We develop a new variational inference algorithm for performing fast approximate inference in the proposed DCM-ARD model based on the DSVI framework proposed by \cite{titsias2014doubly};
\item We empirically show (i) the ability of the the proposed approach to recover the true utility function specifications on semi-artificial choice data, (ii) that DCM-ARD can discover new specifications that outperform previous ones from the literature, and (iii) that the developed DSVI algorithm is able to scale to very large datasets and search spaces. 
\end{itemize}

The remainder of this paper is organized as follows. In the next section, we review the relevant literature for this work. Section~\ref{sec:approach} presents the proposed DCM-ARD model and derives a scalable doubly-stochastic variational inference algorithm for performing fast approximate Bayesian inference on it. The corresponding experimental results are presented in Section~\ref{sec:experiments}. The paper ends with the conclusions (Section~\ref{sec:conclusion}).

\section{Literature review}
\label{sec:literature}

The problem of automatically determining the relevant variables for inclusion in a model has been studied to a significant extent in the supervised machine learning literature under the common title of ``feature selection". When using feature selection techniques, the main premise is that the considered data contain redundant or irrelevant variables, which can therefore be removed without consequent loss of information \citep{dash_feature_1997}. The numerous existing approaches are generally classified as wrapper, filter and embedded methods according to the strategy they employ to search for subsets of variables \citep{guyon2003introduction}. Wrappers use the model of interest to score subsets according to the predictive power they allow to achieve. Despite being computationally intensive, wrappers offer a simple way of addressing the problem: a plethora of methods based on simulated annealing \citep{lin2008parameter, BRUSCO201438}, tabu search \citep{fouskakis2008, PACHECO2009506}, evolutionary algorithms \citep{pal_self-crossover-new_1998, vinterbo1999genetic, soufan_dwfs:_2015} and other combinatorial optimization algorithms have already been applied successfully, both for linear and logistic regressions. In comparison, filter methods are independent of the model under consideration; they use ``proxy'' measures such as correlation or mutual information \citep{xing2001feature, hanchuan_peng_feature_2005, vergara_review_2014} to evaluate single features or subsets. While being less computationally intensive than wrappers, filters usually achieve worse results in terms of prediction power. Finally, embedded methods are characterized by the fact that the selection of variables and the estimation of the model are performed simultaneously, in a single process. A good example of such class of methods is the LASSO, initially proposed by \cite{tibshirani1996regression} and successfully applied both to linear \citep{zhang2008sparsity} and logisitic \citep{huttunen_mind_2013, hossain2014model} regressions. Other existing embedded methods make use of mixed integer optimization \citep{Sato2016} or decision trees \citep{muni_genetic_2006, deng2012feature} to effectively incorporate feature selection as part of the training process. 

In the field of discrete choice analysis, interest has recently emerged for methods that are able to ``mitigate'' the need for presumptive structural assumptions. Two main directions of research are explored in the existing literature: the first substitutes DCMs with machine learning classifiers that do not require any prior knowledge concerning the domain \citep{paredes_machine_2017, brathwaite_machine_2017, lheritier_airline_2018, sifringer_enhancing_2018}, while the second focuses on automatizing the utility specification of DCMs by means of data-driven feature selection algorithms \citep{tutz2015variable, PAZ201950, ortelli2019}. 

A particularly elegant class of methods for performing automatic feature selection in the statistics and machine learning literature relies on the concept of automatic relevance determination (ARD) \citep{tipping2001sparse, mackay1996bayesian, bishop2006pattern}. The idea behind this class of approaches consists in specifying the a-priori uncertainty and infer a-posteriori uncertainty about regression coefficients explicitly and hierarchically in a Bayesian framework. However, unfortunately, Bayesian inference in such hierarchical models quickly becomes intractable, and effective and scalable methods are required in order to perform approximate inference. To that end, \cite{bishop2006pattern} presents a type-II maximum likelihood based on variational inference in a linear regression context, where the hyper-parameters of the hierarchical priors are tuned by maximizing the marginal likelihood of the data. This approach was later extended by \cite{drugowitsch2013variational} to a fully Bayesian approach by further considering a normal inverse-gamma prior over the  hyper-parameters  of the  hierarchical  priors, and then performing variational inference to determine the corresponding posterior distributions. Furthermore, the author also considers ARD in a binary logistic regression context. The difficulty in the latter stems from the non-conjugacy of the sigmoid, which required the authors to consider an additional model-specific parametric lower bound on the sigmoid as proposed by \cite{jaakkola2000bayesian}, which can raise the computational cost and compromise accuracy. Recently, highly efficient general-purpose black-box variational inference methods have proposed in the literature \citep{ranganath2014black,titsias2014doubly}, which allow for approximating the required expectations using inexpensive Monte Carlo approximations. In particular, \cite{titsias2014doubly} proposed a doubly stochastic variational inference for performing ARD in binary logistic regression. The approach proposed in this paper builds on the work of \cite{titsias2014doubly} to propose an ARD framework for discrete choice models, and to develop a corresponding efficient variational inference algorithm. 

\section{Approach}
\label{sec:approach}

\subsection{Discrete choice models}
\label{sec:dcm}

Following the Random Utility Maximization (RUM) theory, discrete choice models are based on the assumption that each individual $n \in \{1,\dots,N\}$ is a rational decision-maker that aims at maximizing some utility with respect to the choice set $\C_n$ that is presented to her. A key step in discrete choice modeling is then to specify a function $U_{in}$ that is able to capture the utility of each alternative $i$ for each individual $n$. The utility function is further assumed to be partitioned intro two components: a systematic (or deterministic) utility $V_{in}$ and a random component $\epsilon_{in}$:
\begin{align}
U_{in} = V_{in} + \epsilon_{in},
\end{align}
where $\epsilon_{in}$ is an \mbox{i.i.d.} term that captures the uncertainty stemming from the impossibility of $V_{in}$ to fully capture the choice context. As for the systematic component $V_{in}$, it is typically assumed to be a linear function of the observable explanatory variables $\bx_{in} = \{x_{din}\}_{d=1}^{D_i}$ of the utility of alternative $i$ for each individual $n$ (\mbox{e.g.} alternative characteristics, individual's socio-demographic attributes, etc.):
\begin{align}
V_{in} = \bs\beta_i^\tr \bx_{in} = \sum_{d=1}^{D_i} \beta_{di} \, x_{din},
\label{eq:utility_spec}
\end{align}
where $\bs\beta_i$ is a vector of alternative-specific preference parameters. This accounts for the more general setting where preference parameters may vary between different alternatives. Following the same reasoning, our specification further allows for a variable number of explanatory variables $D_i$ per alternative $i$.

Under the standard multinomial logit assumption that $\epsilon_{in} \sim \mbox{EV}(0,1)$, the probability of individual $n$ selecting alternative $i$ is given by
\begin{align}
P_n(i) = \frac{e^{V_{in}}}{\sum_{j \in \C_n} e^{V_{jn}}}.
\label{eq:likelihood}
\end{align}
Given a dataset of observed choices and corresponding explanatory variables for a population of size $N$, the modeler's objective is to determine the preference parameters $\bs\beta$, which are typically estimated by maximizing the log-likelihood function:
\begin{align}
\bs\beta^* = \arg\max_{\bs\beta} \, \sum_{n=1}^N \sum_{i \in \C_n} y_{in} \log P_n(i),
\end{align}
where $y_{in}$ is a one-hot encoding of the observed choice for the $n^{th}$ individual (\mbox{i.e.} $y_{in}$ takes the value 1 if the individual $n$ chose the alternative $i$, and 0 otherwise), and $\by$ and $\bs\beta$ are used to denote the set of all observed choices and preference parameters, respectively. 

Despite the appealing simplicity of maximum likelihood estimation methods, in this paper we shall follow a Bayesian approach. The latter not only allows us to infer full posterior distributions for the preference parameters $\bs\beta$ that provide for a principled way of performing hypotheses testing \citep{song2017bayesian} and uncertainty quantification, but also enable online learning approaches in which the posterior over the parameters is continuously updated as more data becomes available \citep{danaf2017personalized}. Moreover, most importantly, it will support the development of the automatic utility function specification approach based on ARD proposed in Section~\ref{sec:ard}. 

We begin by introducing the standard Bayesian framework for the discrete choice model specified above, which will serve as the starting point for the proposed approach in Section~\ref{sec:ard}. To enable the Bayesian treatment of model above, we start by placing a prior distribution over the preference parameters for each of the alternatives:
\begin{align}
\bs\beta_i \sim \N(\bs\beta_i | \textbf{0}, \lambda \textbf{I}),
\label{eq:prior}
\end{align}
where $\textbf{I}$ denotes the identity matrix, thus making $\lambda \textbf{I}$ a diagonal covariance matrix parametrized by $\lambda$. 

In order to summarize the entire model, we present below its generative process - a compact description of the model's assumptions regarding how the observed data was generated. 

\begin{enumerate}[(1)]
	\item For each alternative $i$ in the entire choice set $\C$
	\begin{enumerate}[(a)]
		\item Draw preference parameters $\bs\beta_i \sim \N(\bs\beta_i | \textbf{0}, \lambda \textbf{I})$
	\end{enumerate}
	\item For each individual $n \in \{1,\dots,N\}$
	\begin{enumerate}[(a)]
		\item Draw observed choice variable $y_n \sim \mbox{Categorical}(y_n | P_n)$
	\end{enumerate}
\end{enumerate}
The joint probability distribution is then given by
\begin{align}
p(\by, \bs\beta|\lambda) = \Bigg( \prod_{i \in \C} \N(\bs\beta_i | \textbf{0}, \lambda \textbf{I}) \Bigg) \prod_{n=1}^N \prod_{i \in \C_n} (P_n(i))^{y_{in}},
\end{align} 
where we purposely omitted the explicit dependency on the explanatory variables $\bx$ to avoid cluttering the notation. 
Making use of Bayes' theorem, the posterior distribution over the preference parameters $\bs\beta$ is
\begin{align}
p(\bs\beta|\by,\lambda) = \frac{p(\bs\beta|\lambda) \, \prod_{n=1}^N \prod_{i \in \C_n} (P_n(i))^{y_{in}}}{\int p(\bs\beta|\lambda) \prod_{n=1}^N \prod_{i \in \C_n} (P_n(i))^{y_{in}} \, d\bs\beta}.
\label{eq:bayes}
\end{align}
However, the non-conjugacy between the prior (\ref{eq:prior}) and the softmax likelihood in ($\ref{eq:likelihood}$) deems the integral in the denominator intractable, thus making exact inference infeasible. Fortunately, over recent years, we have observed very significative improvements in the accuracy and scalability of approximate Bayesian inference methods, which we shall exploit in Section~\ref{sec:dsvi}.

\subsection{Automatic utility function specification}
\label{sec:ard}

The main of focus of this paper is on leveraging the Bayesian framework and the concept automatic relevance determination (ARD) \citep{tipping2001sparse} to lift the burden of manually searching for an optimal utility function specification for a given discrete choice problem from the modeler. Namely, we wish to automatically determine the relevant variables for the utility function of each alternative $i$, while considering also for different non-linear transformations (\mbox{e.g.} log-transforms, Box-Cox transforms), different continuous variable discretizations, interactions between variables, etc. In order to allow for some of these modeling options and, in particular, variable interactions, let us begin by considering a more flexible parameterization of the utility function in (\ref{eq:utility_spec}). Letting $s_n$ be a categorical socio-economic variable with $K$ categories associated with individual $n$ (\mbox{e.g.} age, income, education or profession), we can allow for interactions with the remaining variables by introducing an unknown parameter per category $\beta_1,\dots,\beta_K$ and defining the utility function for an alternative $i$ as
\begin{align}
V_{in} = \sum_{d=1}^{D_i} \sum_{k=1}^{K_d} \beta_{kdi} \, \delta_k(s_n) \, h(x_{din}) , 
\label{eq:utility_spec_general}
\end{align}
where $\delta_k(s_n)$ is an indicator function, which takes the value 1 if the $n^{th}$ individual belongs to category $k$ and 0 otherwise, and $h(\cdot)$ is an arbitrary function (\mbox{e.g.} logarithm for a log-transform). Kindly notice that the utility specification in (\ref{eq:utility_spec}) is a special case of (\ref{eq:utility_spec_general}), when $K_d = 1$ and $h(\cdot)$ is the identity function. Similarly, this specification also contains one-hot encodings and discretizations of a variable $d$ as special cases by adapting the functions $\delta_k(\cdot)$ and $h(\cdot)$ accordingly. 

Based on (\ref{eq:utility_spec_general}), the problem of automatic utility function specification can then be defined as determining the subset of input dimensions $\mathcal{S}_i \subseteq \{1,\dots,D_i\}$ that best models the observed choices according to a dataset of observed choices, where $\{1,\dots,D_i\}$ is a very large set of possible variable transformations and representations whose usefulness to the model we wish to test. For example, for a cost variable, a modeler may consider including in $\{1,\dots,D_i\}$ the variable itself, its log-transformed value, cost interacted with gender, cost interacted with age, cost interacted with both gender and age, a piecewise linear transformation, etc. The goal is then to determine which subset $\mathcal{S}_i$ of these should be included in the utility function specification $V_i$. 

\begin{figure}[t!]
\begin{center}
	\includegraphics[scale=0.15,trim={0 0cm 0 0cm},clip]{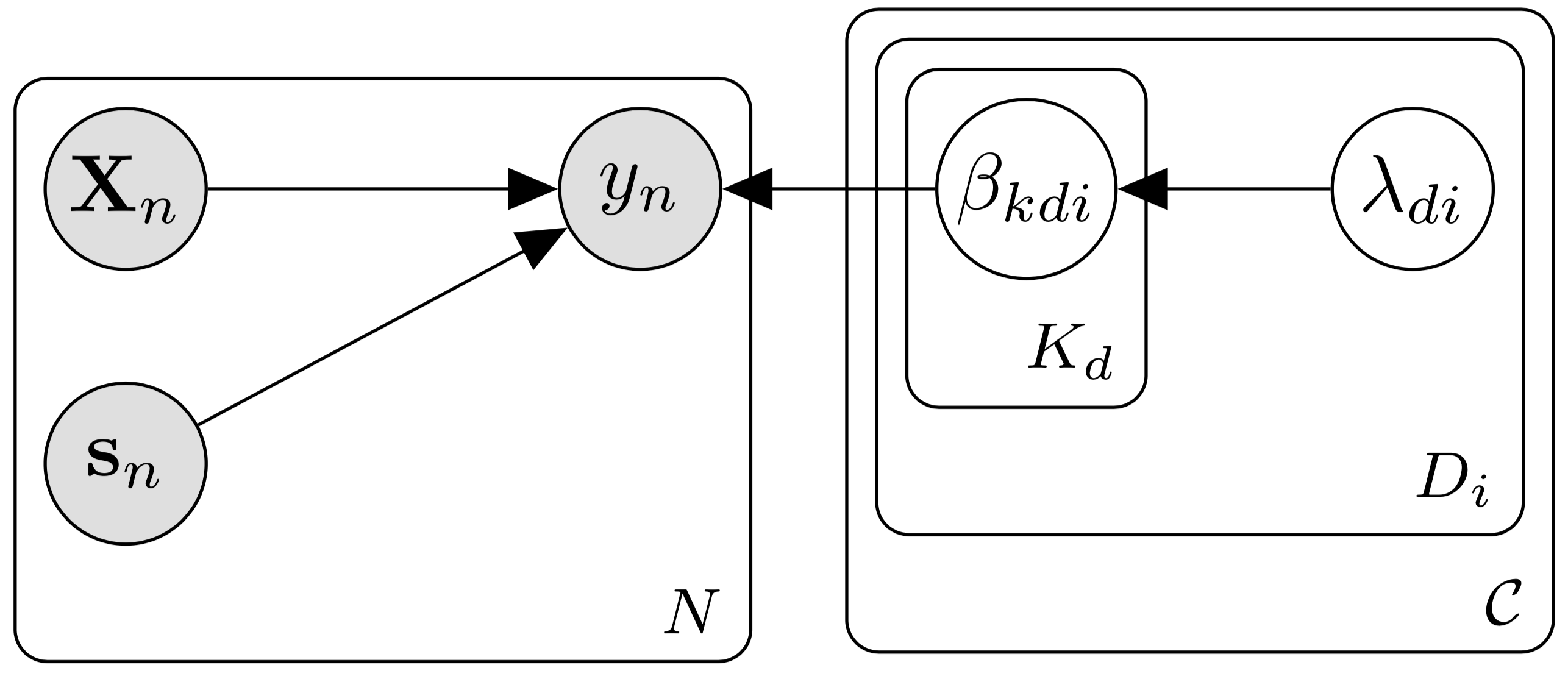}
	\caption{Graphical model representation of the proposed model. Nodes and arrows denote variables and dependencies, respectively. Rectangular plates are used to denote repetition of the structure enclosed within them a given number of times. }
	\label{fig:pgm}
\end{center}
\end{figure}

The starting point for our proposed approach is the concept of automatic relevance determination (ARD), as used for instance in the statistical machine learning literature for the relevance vector machine \citep{tipping2001sparse}. The key idea lies in realizing that preference parameters of irrelevant dimensions $d$ should be pushed towards zero. However, the standard prior specification in (\ref{eq:prior}) is too restrictive to allow for some parameters to be pushed arbitrarily close to zero, while others retain their actual values. This restriction stems for the fact in (\ref{eq:prior}), the parameters are assumed to have independent univariate Gaussian priors that share the same prior variance $\lambda$. Therefore, we can make progress towards ARD in discrete choice models by constructing a flexible hierarchical prior, in which each parameter is assigned an independent Gaussian prior with its own variance, but parameters belonging to the representation of the same variable share the same variance. Mathematically, this corresponds to
\begin{align}
\beta_{kdi} \sim \N(\beta_{kdi} | 0, \lambda_{di}).
\label{eq:beta_kdi_prior}
\end{align}
Please note that the constraint of sharing the same variance over the index $k$ is crucial in order to ensure that the entire group is treated as a whole, \mbox{i.e.} either all $k$ ``sub-dimensions" of a variable $d$ are deemed relevant by the model, or none is and their corresponding parameters are all pushed towards zero. The prior over all the preference parameters is then given by
\begin{align}
p(\bs\beta|\bs\lambda) = \Bigg( \prod_{i \in \C} \prod_{d=1}^{D_i} \prod_{k=1}^{K_d} \N(\beta_{kdi} | 0, \lambda_{di}) \Bigg),
\end{align}
where $\bs\lambda$ is used to denote the set of all $\lambda_{di}$. While one could further place a Gamma prior over the precisions $\lambda_{di}^{-1}$, we refrain from doing so because (i) it would introduce a new set of hyper-parameters to specify and (ii), as we shall see in Section~\ref{sec:dsvi}, it is possible to optimize over the variance parameters $\lambda$ analytically. Hence, we shall continue by treating the latter as point parameters rather than random variables in a fully Bayesian setting. The generative process of the proposed model can then be summarized as follows: 
\begin{enumerate}[(1)]
	\item For each alternative $i$ in the entire choice set $\C$
	\begin{enumerate}[(a)]
		\item For each variable $d \in \{1,\dots,D_i\}$
		\begin{enumerate}[(i)]
			\item Set preference parameter variance $\lambda_{di}$
			\item For each category $k \in \{1,\dots,K_d\}$
			\begin{enumerate}[(a)]
				\item Draw preference parameter $\beta_{kdi} \sim \N(\beta_{kdi} | 0, \lambda_{di})$
			\end{enumerate}
		\end{enumerate}
	\end{enumerate}
	\item For each individual $n \in \{1,\dots,N\}$
	\begin{enumerate}[(a)]
		\item Draw observed choice variable $y_n \sim \mbox{Categorical}(y_n | P_n)$
	\end{enumerate}
\end{enumerate}
In order to place further emphasis on the hierarchical structure of the proposed model, Figure~\ref{fig:pgm} shows a graphical model representation, which highlights the dependencies between the different variables.  

Based on the model specification above, our goal is to be able to jointly infer the preference parameters $\bs\beta$ and estimate the variance parameters $\lambda_{di}$ for each explanatory variable, in order to assess which ones should be included in each utility function $V_i$. As for the ``standard" discrete choice model in Section~\ref{sec:dcm}, performing exact Bayesian inference in the proposed model is intractable. Therefore, we shall proceed by developing an approximate Bayesian inference algorithm using doubly stochastic variational inference \citep{titsias2014doubly}. 

\subsection{Doubly stochastic variational inference}
\label{sec:dsvi}

The intractability of exact inference for the proposed model stems from the impossibility of obtaining an analytical expression for the marginal likelihood in the denominator of (\ref{eq:bayes}), which for the proposed ARD model takes the form
\begin{align}
p(\by|\bs\lambda) = \int \Bigg( \prod_{i \in \C} \prod_{d=1}^{D_i} \prod_{k=1}^{K_d} \N(\beta_{kdi} | 0, \lambda_{di}) \Bigg) \prod_{n=1}^N \prod_{i \in \C_n} (P_n(i))^{y_{in}} \, d\bs\beta.
\label{eq:marginal_lik}
\end{align}
In order to obtain an efficient and scalable approximate inference algorithm that is able to cope with large datasets and with very high dimensionalities $D_i$, we propose the use of the variational inference framework \citep{jordan1999introduction}. 

Variational inference, or variational Bayes, constructs an approximation to the true posterior distribution $p(\bs\beta|\by)$ by considering a family of tractable distributions $q(\bs\beta)$, which can be obtained by relaxing some constraints in the true distribution. In this case, we shall assume the variational distribution $q(\bs\beta)$ to be a fully-factorized (mean-field) approximation to the true posterior:
\begin{align}
\label{eq:approx_q}
q(\bs\beta|\bs\mu,\textbf{c}) = \prod_{i \in \C} \prod_{d=1}^{D_i} \prod_{k=1}^{K_d} \N(\beta_{kdi} | \mu_{kdi}, c_{kdi}),
\end{align}
with variational parameters $\bs\mu$ and $\textbf{c}$. 
The inference problem is then to find the parameters of the variational distribution so that the approximation becomes as close as possible to the true posterior, thereby reducing inference to an optimization problem.

The closeness between the approximate posterior $q(\bs\beta|\bs\mu,\textbf{c})$ and the true posterior $p(\bs\beta|\by)$ can be measured by the Kullback-Leibler (KL) divergence \citep{mackay2003information} given by
\begin{align}
\mathbb{KL}(q(\bs\beta|\bs\mu,\textbf{c})||p(\bs\beta|\by)) = \int q(\bs\beta|\bs\mu,\textbf{c}) \log \frac{q(\bs\beta|\bs\mu,\textbf{c})}{p(\bs\beta|\by)} d\bs\beta.
\end{align}
Although the KL cannot be minimized directly, following the theory on variational inference \citep{jordan1999introduction,mackay2003information}, the KL minimization can be equivalently formulated as maximizing the following lower bound on the log marginal likelihood (or log evidence) in (\ref{eq:marginal_lik}):
\begin{align}
\log p(\by|\bs\lambda) &= \log \int q(\bs\beta|\bs\mu,\textbf{c}) \, \frac{p(\by,\bs\beta|\bs\lambda)}{q(\bs\beta|\bs\mu,\textbf{c})} \, d\bs\beta\nonumber\\
&\geq \int q(\bs\beta|\bs\mu,\textbf{c}) \log \frac{p(\by,\bs\beta|\bs\lambda)}{q(\bs\beta|\bs\mu,\textbf{c})} \, d\bs\beta\nonumber\\
&= \E_{q(\bs\beta)}[\log p(\by,\bs\beta|\bs\lambda)] - \E_{q(\bs\beta)}[\log q(\bs\beta|\bs\mu,\textbf{c})] = \mathcal{L}(\bs\mu,\textbf{c},\bs\lambda),
\label{eq:lowerbound1}
\end{align}
where we made use of Jensen's inequality. We can further write the evidence lower bound, $\mathcal{L}(\bs\mu,\textbf{c},\bs\lambda)$, as a function of simpler terms by exploiting the factorization of the joint and prior distributions, yielding 
\begin{align}
\mathcal{L}(\bs\mu,\textbf{c},\bs\lambda) &= \sum_{i \in \C} \sum_{d=1}^{D_i} \sum_{k=1}^{K_d} \E_{q(\bs\beta)}[\log \N(\beta_{kdi} | 0, \lambda_{di})] + \sum_{n=1}^N \sum_{i \in \C_n} y_{in} \E_{q(\bs\beta)}[\log P_n(i)] \nonumber\\
&- \sum_{i \in \C} \sum_{d=1}^{D_i} \sum_{k=1}^{K_d} \E_{q(\bs\beta)}[\log \N(\beta_{kdi} | \mu_{kdi}, c_{kdi})]
\label{eq:lowerbound2}
\end{align}

Our goal is then to find the variational parameters $\{\bs\mu,\textbf{c}\}$ and the hyper-parameters $\bs\lambda$ that maximize $\mathcal{L}(\bs\mu,\textbf{c},\bs\lambda)$. However, due to the log-sum-exp term resultant from the denominator of the softmax, the expectation $\E_{q(\bs\beta)}[\log P_n(i)]$ in (\ref{eq:lowerbound2}) is still intractable. While some authors proposed the use of computationally expensive approximations to further bound this term \citep{blei2007correlated,knowles2011non}, we shall rely on a more efficient and scalable approximation based on the theory of stochastic optimization. In order to enable it, we begin by reparameterizing our approximate distribution in (\ref{eq:approx_q}). 

Consider a random variable $z \sim \N(z|0,1)$. We can change the mean and variance by applying an invertible transformation $\beta = c z + \mu$ and making use of the change of variables formula for a random vector, which states that for a given function $f(x)$, and given an invertible transformation $y = h(x)$, we have that $f(y) = f(h(x)) |J_{h^{-1}}|$, where $|J_{h^{-1}}|$ denotes the determinant of the Jacobian matrix of the inverse transformation $h^{-1}$. Hence, given the transformation $\beta = c z + \mu$ and its inverse $z = c^{-1} (\beta - \mu)$, we can rewrite the approximate distribution in (\ref{eq:approx_q}) as 
\begin{align}
q(\bs\beta|\bs\mu,\textbf{c}) &= \prod_{i \in \C} \prod_{d=1}^{D_i} \prod_{k=1}^{K_d} \frac{1}{|c_{kdi}|} \N(c_{kdi}^{-1} (\beta_{kdi} - \mu_{kdi})|0,1).
\label{eq:q_w_l}
\end{align}
By plugging (\ref{eq:q_w_l}) into (\ref{eq:lowerbound1}) and changing variables according to 
$z = c^{-1} (\beta - \mu)$, we can rewrite $\mathcal{L}(\bs\mu,\textbf{c},\bs\lambda)$ as follows:
\begin{align}
\mathcal{L}(\bs\mu,\textbf{c},\bs\lambda) &= \int \N(\bz|\textbf{0},\textbf{I}) \log \frac{p(\by,\textbf{c} \circ \textbf{z} + \bs\mu|\bs\lambda) \prod_{i \in \C} \prod_{d=1}^{D_i} \prod_{k=1}^{K_d} |c_{kdi}|}{\N(\bz|\textbf{0},\textbf{I})} \, d\bz\nonumber\\
&= \E_{\N(\bz|\textbf{0},\textbf{I})}[\log p(\by|\textbf{c} \circ \textbf{z} + \bs\mu)]  + \sum_{i \in \C} \sum_{d=1}^{D_i} \sum_{k=1}^{K_d} \log c_{kdi}\nonumber\\
&+ \sum_{i \in \C} \sum_{d=1}^{D_i} \sum_{k=1}^{K_d}  \E_{\N(z_{kdi}|0,1)} \log \N(c_{kdi} z_{kdi} + \mu_{kdi}|0,\lambda_{di}) + \mbox{const.},
\label{eq:lowerbound3}
\end{align}
where $\circ$ is used to denote the element-wise product and we used the factorization of the joint distribution $p(\by,\textbf{c} \circ \textbf{z} + \bs\mu|\bs\lambda)$ in the last step. The term $-\E_{\N(\bz|\textbf{0},\textbf{I})}[\log \N(\bz|\textbf{0},\textbf{I})]$ was ignored because it is constant \wrt the variational parameters. Making use of the Gaussian pdf and linearity of expectation leads to the final evidence lower bound
\begin{align}
\mathcal{L}(\bs\mu,\textbf{c},\bs\lambda) &= \E_{\N(\bz|\textbf{0},\textbf{I})}[\log p(\by|\textbf{c} \circ \textbf{z} + \bs\mu)]  + \sum_{i \in \C} \sum_{d=1}^{D_i} \sum_{k=1}^{K_d} \log c_{kdi}\nonumber\\
&- \frac{1}{2} \sum_{i \in \C} \sum_{d=1}^{D_i} K_d \log \lambda_{di} - \frac{1}{2} \sum_{i \in \C} \sum_{d=1}^{D_i} \sum_{k=1}^{K_d} \frac{c_{kdi}^2 + \mu_{kdi}^2}{\lambda_{di}} + \mbox{const.}
\label{eq:lowerbound4}
\end{align}
The key insight is that, through the change of variables, the variational parameters have been transferred inside the log likelihood, thus enabling stochastic optimization by sampling gradients from it. 

Regarding the variance hyper-parameters $\bs\lambda$, as it turns out, it is possible to optimize them analytically. This contrasts with other applications of ARD, where the prior variances are estimated using Expectation-Maximization (EM) - a procedure that can exhibit slow convergence due to the strong dependency between the variational parameters $\{\bs\mu,\textbf{c}\}$ and the hyper-parameters $\bs\lambda$ \citep{titsias2014doubly}. Taking derivatives of (\ref{eq:lowerbound4}) \wrt $\lambda_{di}$ and setting them to zero yields the following optimum:
\begin{align}
\lambda_{di}^* = \frac{1}{K_d} \sum_{k=1}^{K_d} (c_{kdi}^2 + \mu_{kdi}^2).
\label{eq:lambda}
\end{align}
Substituting back these optimal values in $\mathcal{L}(\bs\mu,\textbf{c},\bs\lambda)$ gives the optimized evidence lower bound
\begin{align}
\mathcal{L}(\bs\mu,\textbf{c}) &= \E_{\N(\bz|\textbf{0},\textbf{I})}[\log p(\by|\textbf{c} \circ \textbf{z} + \bs\mu)]  + \sum_{i \in \C} \sum_{d=1}^{D_i} \sum_{k=1}^{K_d} \log c_{kdi} - \frac{1}{2} \sum_{i \in \C} \sum_{d=1}^{D_i} K_d \log \sum_{k=1}^{K_d} (c_{kdi}^2 + \mu_{kdi}^2).
\label{eq:lowerbound5}
\end{align}

In order to fit the variational distribution to the true posterior, we must optimize the lower bound in (\ref{eq:lowerbound5}) \wrt $\bs\mu$ and $\textbf{c}$. Taking derivatives gives:
\begin{align}
\nabla_{\mu_{kdi}} \mathcal{L}(\bs\mu,\textbf{c}) &= \E_{\N(\bz|\textbf{0},\textbf{I})}[\nabla_{\mu_{kdi}} \log p(\by|\textbf{c} \circ \textbf{z} + \bs\mu)] - K_d \frac{\mu_{kdi}}{\sum_{k=1}^{K_d} c_{kdi}^2 + \mu_{kdi}^2}\nonumber\\
\nabla_{c_{kdi}} \mathcal{L}(\bs\mu,\textbf{c}) &= \E_{\N(\bz|\textbf{0},\textbf{I})}[\nabla_{c_{kdi}} \log p(\by|\textbf{c} \circ \textbf{z} + \bs\mu)] + \frac{1}{c_{kdi}} - K_d \frac{c_{kdi}}{\sum_{k=1}^{K_d} c_{kdi}^2 + \mu_{kdi}^2}\nonumber
\end{align}
We can further rewrite these derivatives by changing variables in the reverse direction, $\beta = cz + \mu$, and making use of the chain rule, thus leading to the final gradients:
\begin{align}
\nabla_{\mu_{kdi}} \mathcal{L}(\bs\mu,\textbf{c}) &= \E_{\N(\bs\beta|\bs\mu,\textbf{c})}[\nabla_{\beta_{kdi}} \log p(\by|\bs\beta)] - K_d \frac{\mu_{kdi}}{\sum_{k=1}^{K_d} c_{kdi}^2 + \mu_{kdi}^2}\nonumber\\
\nabla_{c_{kdi}} \mathcal{L}(\bs\mu,\textbf{c}) &= \E_{\N(\bs\beta|\bs\mu,\textbf{c})}[\nabla_{\beta_{kdi}} \log p(\by|\bs\beta)] \times c_{kdi}^{-1} (\beta_{kdi} - \mu_{kdi}) + \frac{1}{c_{kdi}} - K_d \frac{c_{kdi}}{\sum_{k=1}^{K_d} c_{kdi}^2 + \mu_{kdi}^2}\nonumber
\end{align}
As for the gradients of the log likelihood of the discrete choice model specified in Section~\ref{sec:ard}, they are given by
\begin{align}
\nabla_{\beta_{kdi}} \log p(\by|\bs\beta) 
&= \sum_{n=1}^N y_{in} \delta_k(s_n) \, h(x_{din}) - \sum_{n=1}^N \delta_k(s_n) \, h(x_{din}) P_n(i).
\end{align}

The lower bound $\mathcal{L}(\bs\mu,\textbf{c})$ can then be optimized by first sampling a set of preference parameters $\bs\beta = \textbf{c} \circ \textbf{z} + \bs\mu$, $\textbf{z} \sim \N(\textbf{0},\textbf{I})$, and using the stochastic gradients above to update the all variational parameters $\bs\mu$ and $\textbf{c}$ in parallel:
\begin{align}
\bs\mu^{(t)} = \bs\mu^{(t-1)} + \rho_t \nabla_{\bs\mu} \mathcal{L}(\bs\mu,\textbf{c}) \\
\textbf{c}^{(t)} = \textbf{c}^{(t-1)} + \rho_t \nabla_{\textbf{c}} \mathcal{L}(\bs\mu,\textbf{c})
\end{align}
Following the theory of stochastic optimization \citep{robbins1985stochastic}, using a schedule of the learning rates $\{\rho_t\}$ such that $\sum \rho_t = \infty$,
$\sum \rho_t^2 < \infty$, the iteration in Algorithm 1 will converge to a local maxima of the bound in (\ref{eq:lowerbound5}) or to the global maximum when this bound is concave. At convergence, we can assess the relevancy of each explanatory variable $d$ in the utility function for alternative $i$ by evaluating the magnitude of the estimated variance parameter $\lambda_{di}$ using (\ref{eq:lambda}). 

Lastly, we can further scale-up the variational inference algorithm described above by introducing a second type of stochasticity as proposed by \cite{hoffman2013stochastic}. This second type of stochastic stems from using ``mini-batches" of data to compute the stochastic gradients rather then the entire dataset at once, hence resulting in a doubly stochastic variational inference algorithm. The final procedure is summarized in Algorithm~\ref{algorithm}. As we shall see in our experimental results (Section~\ref{sec:experiments}), the proposed inference algorithm is able to scale to very large datasets and perform automatic utility function specification considering a very high number of possible explanatory variables $D_i$. 

\begin{algorithm}[t]
\SetAlgoLined
\KwIn{Set $\bX_i$ of all variables to be tested for the utility function of each alternative $i$ for all individuals in a population; corresponding observed choices $\by$}
\KwOut{Subset of selected variables $\mathcal{S}_i$ for each alternative $i$}
Initialize $\bs\mu^{(0)}$, $\textbf{c}^{(0)}$, t=0\;
 \Repeat{convergence criterion is met}{
  $t = t + 1$\;
  $\bz \sim \N(\textbf{0},\textbf{I})$\;
  $\bs\beta = \textbf{c}^{(t-1)} \circ \textbf{z} + \bs\mu^{(t-1)}$\;
  Sample mini-batch of data to approximate gradients\;
  Update $\bs\mu^{(t)} = \bs\mu^{(t-1)} + \rho_t \nabla_{\bs\mu} \mathcal{L}(\bs\mu,\textbf{c})$\;
  Update $\textbf{c}^{(t)} = \textbf{c}^{(t-1)} + \rho_t \nabla_{\textbf{c}} \mathcal{L}(\bs\mu,\textbf{c})$\;
 }
 Estimate $\bs\lambda^*$ using (\ref{eq:lambda})\;
 \ForAll{i}{
   $\mathcal{S}_i = \{d \in D_i : \lambda_{di}^* \gg 0\}$\;
 }
 \vspace{0.5cm}
 \caption{Automatic utility function specification algorithm.}
 \label{algorithm}
\end{algorithm}

\section{Experiments}
\label{sec:experiments}

In this section, an empirical evaluation of the proposed DCM-ARD for automatic utility function specification is performed based on both semi-artificial and real choice data. For both sets of experiments, the dataset used is the Swissmetro (SM) dataset described in \citep{bierlaire2001acceptance}. This dataset consists of survey data collected on the trains between \mbox{St.} Gallen and Geneva, in which the respondents provided information in order to analyze the impact of the construction of the Swissmetro. The alternatives offered to each respondent were: train, Swissmetro and car (only for car owners). After discarding respondents for which some variables were not available (\mbox{e.g.} age, purpose), a total of 10692 responses from 1188 individuals were used for the experiments. 

The proposed DCM-ARD model and its corresponding doubly-stochastic variational inference (DSVI) algorithm were implemented in Matlab. Source code for the implementation and for reproducing all experiments in this paper is publicly available at: \url{http://fprodrigues.com/DCM-ARD/}. 

\subsection{Semi-artificial choice data}

In order to empirically demonstrate the ability of the proposed approach to discover the correct utility function specifications, we began by conducting an extensive series of experiments on semi-artificial choice data based on the Swissmetro dataset. We manually specified a set of ``artificial'' (but realistic) utility function specifications of varying complexity and, based on the Swissmetro dataset, we sampled new artificial choices for the respondents according to the manually-specified utility functions. This was done by fitting a standard DCM with the manually-specified utility function to the original data using maximum-likelihood estimation and, based on the learned parameters $\bs\beta^*$, we then sampled new choices $y_n \sim \mbox{Categorical}(y_n|P_n)$. 

We consider two experimental settings for the application of DCM-ARD: 
\begin{itemize}
\item an experimental setting with a medium-sized utility function search space, in which the number of possible variables to be included in the utility functions is 252; these include the original variables (\mbox{e.g.} intercept ``ASC", travel-time ``TT", cost ``CO" and headway ``HE"), their log-transformations, and interactions of both the original variables and their logarithms with trip purpose (``pur", 9 categories), respondent age (``age", 5 groups) and annual season ticket availability (``ga", binary). Kindly note that, although this results in 252 variables that can be included in the specification, the dimensionality of the utility function search-space includes all combinations of possible utility functions that can be generated using these variables and therefore grows exponentially with this number. For example, considering just the subset of all utility functions with only 10 variables results in $\binom{252}{10} = 2.4 \times 10^{17}$ possible utility functions to be considered; 
\item an experimental setting with a large utility function search space; besides the variables in the medium-sized search space, this search space also considers Box-Cox transformations, variable segmentations based on K-means clustering, and interactions of the original variables with respondent income (``inc", 5 groups), luggage (``lug": none, one piece or multiple pieces) and who pays for the trip (``who": unknown, self, employer or half-half). This results in a total of 602 possible variables to be included in the utility function specifications. 
\end{itemize}

\begin{table}[t]
\caption{Manually-defined utility function specifications used to generate the semi-artificial choice data.}
\begin{center}
\setlength\tabcolsep{5pt}
\begin{tabular}{c | l | l | l}
& \multicolumn{3}{c}{Artificial specification}\\
\hline
Spec & Variables in $V_{\mbox{\scriptsize train}}$ & Variables in $V_{\mbox{\scriptsize sm}}$ & Variables in $V_{\mbox{\scriptsize car}}$\\
\hline
\hline
\multirow{1}{*}{S1} & ASC, TT, CO & ASC, TT, CO & TT, CO\\ 
\hline
\multirow{2}{*}{S2} & ASC, TT, TT x age, CO & ASC, TT, CO,  & TT, TT x age, \\ 
& & CO x ga & CO \\
\hline
\multirow{2}{*}{S3} & ASC, TT, TT x age,  & ASC, TT, CO, & TT, TT x age, \\ 
& CO, CO x ga, HE &  CO x ga, log(HE) & CO \\
\hline
\hline
\multirow{2}{*}{S4} & ASC, ASC x ga, TT, CO & ASC, ASC x ga,  & TT, \underline{CO}, \\ 
& & TT, CO & CO x purpose \\
\hline
\multirow{1}{*}{S5} & ASC, log(TT), HE & ASC, log(TT), \underline{HE} & TT, CO \\ 
\hline
\multirow{2}{*}{S6} & ASC, log(TT), & ASC, log(TT) & TT, CO \\ 
& log(TT) x ga, CO & & \\
\hline
\hline
\multirow{2}{*}{S7} & ASC, \underline{box(TT)},  & ASC, TT  & TT, CO \\ 
& box(TT) x ga, CO & & \\
\hline
\multirow{2}{*}{S8} & ASC, ASC x ga, TT,  & ASC, ASC x ga, TT, & TT, CO, \\ 
& CO, CO x who & CO, CO x who & \underline{CO x luggage} \\
\hline
\multirow{2}{*}{S9} & ASC, TT, CO & ASC, TT, TT x age, & TT, CO, \\ 
& CO x ga & CO, CO x ga & CO x income\\
\end{tabular}
\end{center}
\label{table:fakespecs}
\end{table}%

Based on these two search spaces, we manually defined 9 artificial utility function specifications as shown in Table~\ref{table:fakespecs}. Specifications S1-S6 are based on the medium-sized search space, while specifications S7-S9 are based on the large search space. However, in order to verify that DCM-ARD is able to discover the true utility function specification used to generate the choice data regardless of how large the search space considered is, we also test specifications S1-S3 with the large search space. \footnote{We further tested other specifications, but omitted their results for conciseness (they lead to similar conclusions). However, they are available at: \url{http://fprodrigues.com/DCM-ARD/}}

\begin{table}[t]
\caption{Results of DCM-ARD for medium-sized search space (part 1).}
\begin{center}
\setlength\tabcolsep{5pt}
\begin{tabular}{c| lc | lc | lc}
& \multicolumn{2}{c|}{Train} & \multicolumn{2}{c|}{Swiss Metro} & \multicolumn{2}{c}{Car}\\
\hline
 Spec & Variable & $\lambda$ & Variable & $\lambda$ & Variable & $\lambda$\\
\hline
\multirow{5}{*}{S1} & \textbf{ASC} & \textbf{1.814} & \textbf{TT} & \textbf{0.513} & \textbf{TT} & \textbf{0.744}\\
& \textbf{TT} & \textbf{1.174} & \textbf{ASC} & \textbf{0.126} & \textbf{CO} & \textbf{0.011}\\
& \textbf{CO} & \textbf{0.393} & \textbf{CO} & \textbf{0.066} & log(TT) x pur1 & 0.000\\
& CO x age1 & 0.000 & log(HE) x age1 & 0.000 & log(TT) x pur2 & 0.000\\
& \multicolumn{2}{c|}{...}& \multicolumn{2}{c|}{...}& \multicolumn{2}{c}{...}\\
\hline
\multirow{9}{*}{S2} & \textbf{ASC} & \textbf{2.353} & \textbf{TT} & \textbf{0.495} & \textbf{TT} & \textbf{0.389}\\
& \textbf{TT x age1} & \textbf{0.524} & \textbf{CO x ga} & \textbf{0.195} & \textbf{CO} & \textbf{0.070}\\
& \textbf{TT x age2} & \textbf{0.524} & \textbf{ASC} & \textbf{0.120} & \textbf{TT x age1} & \textbf{0.060}\\
& \textbf{TT x age3} & \textbf{0.524} & \textbf{CO} & \textbf{0.030} & \textbf{TT x age2} & \textbf{0.060}\\
& \textbf{TT x age4} & \textbf{0.524} & ASC x pur1 & 0.000 & \textbf{TT x age3} & \textbf{0.060}\\
& \textbf{TT} & \textbf{0.468} & ASC x pur2 & 0.000 & \textbf{TT x age4} & \textbf{0.060}\\
& \textbf{CO} & \textbf{0.416} & ASC x pur3 & 0.000 & log(TT) x pur1 & 0.000\\
& TT x pur1 & 0.000 & ASC x pur4 & 0.000 & log(TT) x pur2 & 0.000\\
& \multicolumn{2}{c|}{...}& \multicolumn{2}{c|}{...}& \multicolumn{2}{c}{...}\\
\hline
\multirow{10}{*}{S3} & \textbf{ASC} & \textbf{2.536} & \textbf{TT} & \textbf{0.522} & \textbf{TT} & \textbf{0.478}\\
& \textbf{CO x ga} & \textbf{0.633} & \textbf{CO x ga} & \textbf{0.426} & \textbf{CO} & \textbf{0.120}\\
& \textbf{TT x age1} & \textbf{0.510} & \textbf{ASC} & \textbf{0.133} & \textbf{TT x age1} & \textbf{0.061}\\
& \textbf{TT x age2} & \textbf{0.510} & \textbf{CO} & \textbf{0.023} & \textbf{TT x age2} & \textbf{0.061}\\
& \textbf{TT x age3} & \textbf{0.510} & \textbf{log(HE)} & \textbf{0.005} & \textbf{TT x age3} & \textbf{0.061}\\
& \textbf{TT x age4} & \textbf{0.510} & HE x age1 & 0.000 & \textbf{TT x age4} & \textbf{0.061}\\
& \textbf{TT} & \textbf{0.300} & HE x age2 & 0.000 & log(TT) x ga & 0.000\\
& \textbf{CO} & \textbf{0.202} & HE x age3 & 0.000 & log(CO) x pur1 & 0.000\\
& \textbf{HE} & \textbf{0.056} & HE x age4 & 0.000 & log(CO) x pur2 & 0.000\\
& HE x pur1 & 0.000 & log(CO) x pur1 & 0.000 & log(CO) x pur3 & 0.000\\
& \multicolumn{2}{c|}{...}& \multicolumn{2}{c|}{...}& \multicolumn{2}{c}{...}\\
\end{tabular}
\end{center}
\label{table:fake_results_easy1}
\end{table}%

\begin{table}[t]
\caption{Results of DCM-ARD for medium-sized search space (part 2).}
\begin{center}
\setlength\tabcolsep{5pt}
\begin{tabular}{c| lc | lc | lc}
& \multicolumn{2}{c|}{Train} & \multicolumn{2}{c|}{Swiss Metro} & \multicolumn{2}{c}{Car}\\
\hline
 Spec & Variable & $\lambda$ & Variable & $\lambda$ & Variable & $\lambda$\\
\hline
\multirow{11}{*}{S4} & \textbf{ASC x ga} & \textbf{6.836} & \textbf{ASC x ga} & \textbf{3.401} & \textbf{TT} & \textbf{0.855}\\
& \textbf{CO} & \textbf{2.323} & \textbf{CO} & \textbf{1.354} & \textbf{CO x pur1} & \textbf{0.100}\\
& \textbf{ASC} & \textbf{1.338} & \textbf{TT} & \textbf{0.462} & \textbf{CO x pur2} & \textbf{0.100}\\
& \textbf{TT} & \textbf{0.885} & \textbf{ASC} & \textbf{0.361} & \textbf{CO x pur3} & \textbf{0.100}\\
& CO x ga & 0.001 & log(HE) x age1 & 0.001 & \textbf{CO x pur4} & \textbf{0.100}\\
& ASC x pur1 & 0.000 & log(HE) x age2 & 0.001 & \textbf{CO x pur5} & \textbf{0.100}\\
& ASC x pur2 & 0.000 & log(HE) x age3 & 0.001 & \textbf{CO x pur6} & \textbf{0.100}\\
& ASC x pur3 & 0.000 & log(HE) x age4 & 0.001 & \textbf{CO x pur7} & \textbf{0.100}\\
& ASC x pur4 & 0.000 & CO x pur1 & 0.000 & \textbf{CO x pur8} & \textbf{0.100}\\
& ASC x pur5 & 0.000 & CO x pur2 & 0.000 & log(TT) x ga & 0.000\\
& \multicolumn{2}{c|}{...}& \multicolumn{2}{c|}{...}& \multicolumn{2}{c}{...}\\
\hline
\multirow{6}{*}{S5} & \textbf{ASC} & \textbf{1.775} & \textbf{log(TT)} & \textbf{0.557} & \textbf{TT} & \textbf{0.722}\\
& \textbf{log(TT)} & \textbf{1.405} & \textbf{ASC} & \textbf{0.087} & \textbf{CO} & \textbf{0.042}\\
& \textbf{HE} & \textbf{0.035} & CO & 0.002 & log(TT) x pur1 & 0.000\\
& TT x age1 & 0.000 & \underline{HE} & 0.001 & log(TT) x pur2 & 0.000\\
& TT x age2 & 0.000 & HE x age1 & 0.000 & log(TT) x pur3 & 0.000\\
& \multicolumn{2}{c|}{...}& \multicolumn{2}{c|}{...}& \multicolumn{2}{c}{...}\\
\hline
\multirow{6}{*}{S6} & \textbf{ASC} & \textbf{2.071} & \textbf{log(TT)} & \textbf{0.664} & \textbf{TT} & \textbf{0.809}\\
& \textbf{log(TT) x ga} & \textbf{1.600} & \textbf{ASC} & \textbf{0.106} & \textbf{CO} & \textbf{0.042}\\
& \textbf{log(TT)} & \textbf{0.611} & log(TT) x age1 & 0.000 & log(TT) x pur1 & 0.000\\
& \textbf{CO} & \textbf{0.394} & log(TT) x age2 & 0.000 & log(TT) x pur2 & 0.000\\
& TT x age1 & 0.000 & log(TT) x age3 & 0.000 & log(TT) x pur3 & 0.000\\
& \multicolumn{2}{c|}{...}& \multicolumn{2}{c|}{...}& \multicolumn{2}{c}{...}\\
\end{tabular}
\end{center}
\label{table:fake_results_easy2}
\vspace{-0.2cm}
\end{table}%

Given the semi-artificial choice data generated based on the manually-defined utility function specifications from Table~\ref{table:fakespecs}, our goal is to test the ability DCM-ARD to recover the correct utility function specifications in a purely data-driven way. Tables~\ref{table:fake_results_easy1} and \ref{table:fake_results_easy2} show the top-K variables selected by DCM-ARD for the medium-sized search space (\mbox{i.e.} specifications S1-S6) ranked according to their respective learned $\lambda$ values. In order to simplify the analysis of the results, the variables deemed relevant by DCM-ARD are highlighted in bold. Irrelevant variables are expected to have $\lambda \approx 0$.  As these results demonstrate, the proposed DCM-ARD is able to discover the true specifications almost perfectly, with all the truly ``irrelevant" variables being assigned a $\lambda$ value of approximately zero. The only minor exceptions can be found in specifications S4 and S5. In the learned utility function for S4, we can observe that cost (``CO") is assigned a $\lambda$ value of zero for the utility of car despite the fact that it was part of the true specification that was used to generate the semi-artificial data. We believe this to be a consequence of the inclusion of the interaction between ``CO" and purpose (``pur") in the true specification for car. Since there is a total of 9 different purposes and some of them have an extremely low number of observations, the effect of ``CO" alone can be captured by the baseline and therefore its presence in the specification is essentially not required from a pure data perspective. As for S5, the headway variable (``HE") in the SM utility was assigned a rather low value of $\lambda$ ($\lambda = 0.001$), despite the fact that it should be clearly identified by DCM-ARD as a relevant variable, since it was part of the true specification of S5. 

\begin{figure*}[t!]
\begin{center}
\subfloat[Evidence lower bound $\mathcal{L}(\bs\mu,\textbf{c})$]{\includegraphics[scale=0.33,trim={0 0cm 0 0cm},clip]{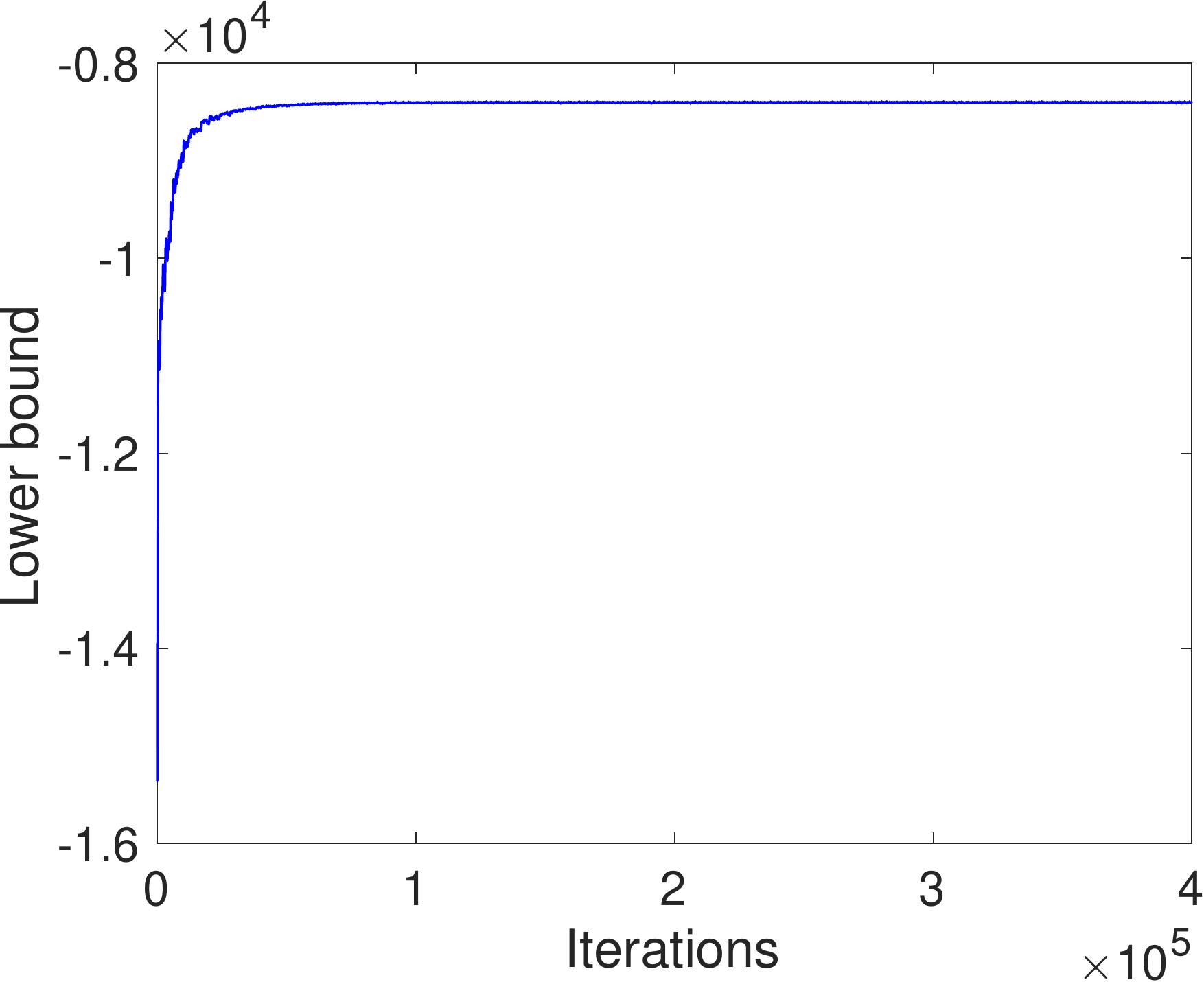} \label{subfig:elbo}} \hspace{1cm}
\subfloat[Learned $\lambda$ values]{\includegraphics[scale=0.33,trim={0 0cm 0 0cm},clip]{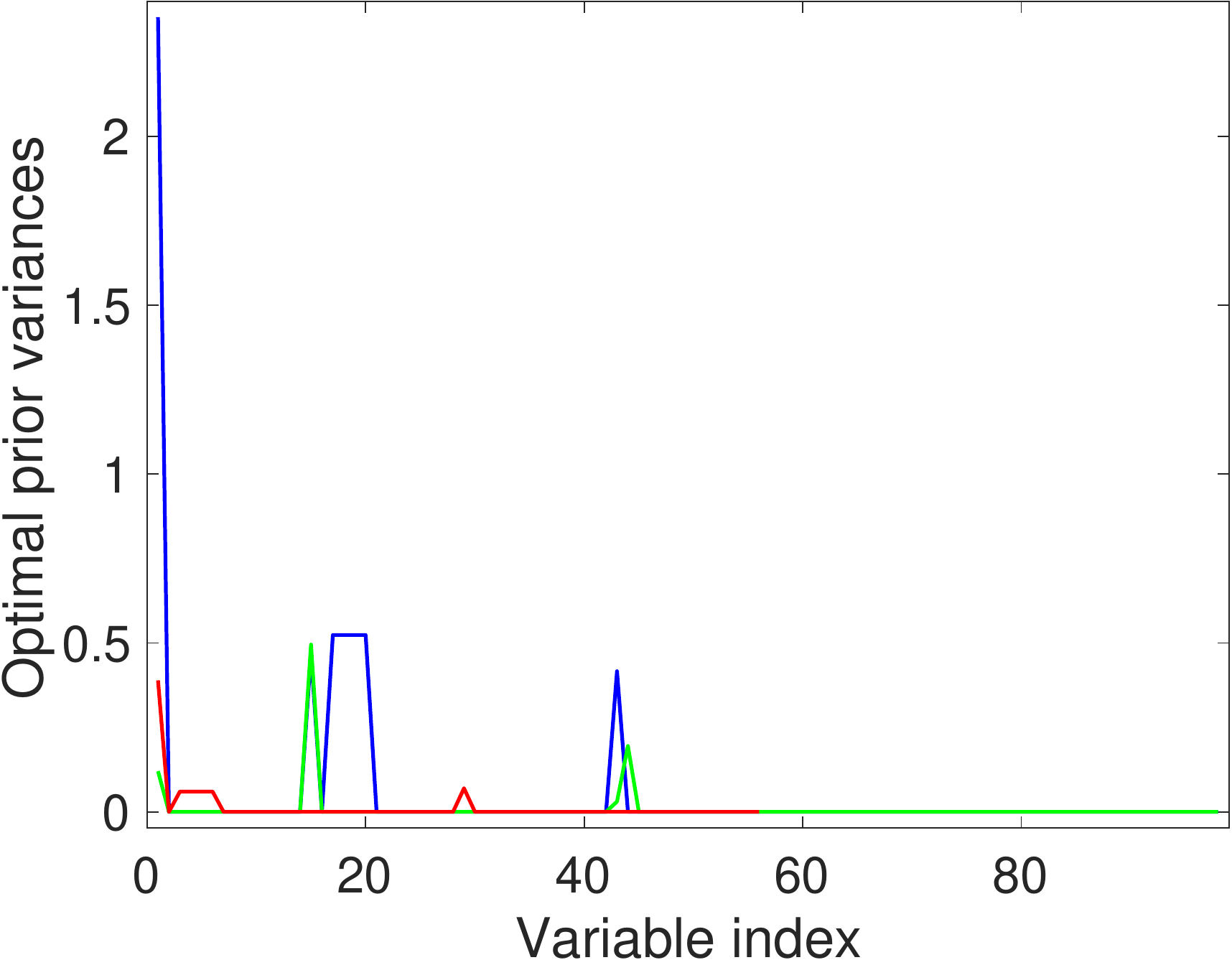} \label{subfig:lambda}}
\caption{Convergence of the evidence lower bound $\mathcal{L}(\bs\mu,\textbf{c})$ and learned $\lambda$ values for the different utility functions (blue: Train, green: SM, red: Car) in specification S2.}
\label{fig:elbo_lambda}
\vspace{-0.2cm}
\end{center}
\end{figure*}

In order to provide a deeper understanding of the proposed approach for automatic utility function specification, Figure~\ref{subfig:elbo} shows the convergence of the derived DSVI algorithm when applied for specification S2 and Figure~\ref{subfig:lambda} gives a broader perspective on the sparsity induced by the hierarchical prior that DCM-ARD uses. While Figure~\ref{subfig:elbo} demonstrates that the proposed DSVI algorithm is able to converge within a few thousand iterations (mini-batches), Figure~\ref{subfig:lambda} illustrates that the learned optimal prior variances $\lambda$ for the S2 semi-artificial choice data are zero for the majority of the input dimensions, except for the few dimensions that correspond to variables that actually belong to the true utility function specification (S2) that was used to generate the data. Furthermore, one can observe two non-zero ``plateaus" (one blue and one red) that correspond to the $\lambda$ values of the interacted variables in S2, which are enforced by the DCM-ARD model to be considered jointly through the tying of the variance parameters of the Gaussian priors (see Eq.~\ref{eq:beta_kdi_prior}).

\begin{table}[t]
\caption{Results of DCM-ARD for large search space (part 1).}
\begin{center}
\setlength\tabcolsep{5pt}
\begin{tabular}{c| lc | lc | lc}
& \multicolumn{2}{c|}{Train} & \multicolumn{2}{c|}{Swiss Metro} & \multicolumn{2}{c}{Car}\\
\hline
 Spec & Variable & $\lambda$ & Variable & $\lambda$ & Variable & $\lambda$\\
\hline
\multirow{5}{*}{S1} & \textbf{ASC} & \textbf{1.879} & \textbf{TT} & \textbf{0.537} & \textbf{TT} & \textbf{0.677}\\
& \textbf{TT} & \textbf{1.196} & \textbf{ASC} & \textbf{0.137} & \textbf{CO} & \textbf{0.011}\\
& \textbf{CO} & \textbf{0.513} & \textbf{CO} & \textbf{0.096} & CO x pur1 & 0.000\\
& log(CO) x inc1 & 0.001 & TT x lugg1 & 0.000 & CO x pur2 & 0.000\\
& \multicolumn{2}{c|}{...}& \multicolumn{2}{c|}{...}& \multicolumn{2}{c}{...}\\
\hline
\multirow{9}{*}{S2} & \textbf{ASC} & \textbf{2.391} & \textbf{TT} & \textbf{0.568} & \textbf{TT} & \textbf{0.411}\\
& \textbf{TT} & \textbf{0.606} & \textbf{CO x ga} & \textbf{0.179} & \textbf{TT x age1} & \textbf{0.059}\\
& \textbf{CO} & \textbf{0.477} & \textbf{ASC} & \textbf{0.130} & \textbf{TT x age2} & \textbf{0.059}\\
& \textbf{TT x age1} & \textbf{0.352} & \textbf{CO} & \textbf{0.031} & \textbf{TT x age3} & \textbf{0.059}\\
& \textbf{TT x age2} & \textbf{0.352} & HE x inc1 & 0.000 & \textbf{TT x age4} & \textbf{0.059}\\
& \textbf{TT x age3} & \textbf{0.352} & HE x inc2 & 0.000 & \textbf{CO} & \textbf{0.036}\\
& \textbf{TT x age4} & \textbf{0.352} & HE x inc3 & 0.000 & TT x lugg1 & 0.000\\
& log(CO) x pur1 & 0.001 & HE x inc4 & 0.000 & TT x lugg2 & 0.000\\
& \multicolumn{2}{c|}{...}& \multicolumn{2}{c|}{...}& \multicolumn{2}{c}{...}\\
\hline
\multirow{11}{*}{S3} & \textbf{ASC} & \textbf{2.599} & \textbf{TT} & \textbf{0.594} & \textbf{TT} & \textbf{0.446}\\
& \textbf{CO x ga} & \textbf{0.717} & \textbf{CO x ga} & \textbf{0.477} & \textbf{CO} & \textbf{0.107}\\
& \textbf{TT} & \textbf{0.432} & \textbf{ASC} & \textbf{0.127} & \textbf{TT x age1} & \textbf{0.074}\\
& \textbf{TT x age1} & \textbf{0.364} & \textbf{CO} & \textbf{0.017} & \textbf{TT x age2} & \textbf{0.074}\\
& \textbf{TT x age2} & \textbf{0.364} & \textbf{log(HE)} & \textbf{0.003} & \textbf{TT x age3} & \textbf{0.074}\\
& \textbf{TT x age3} & \textbf{0.364} & HE x inc1 & 0.000 & \textbf{TT x age4} & \textbf{0.074}\\
& \textbf{TT x age4} & \textbf{0.364} & HE x inc2 & 0.000 & CO x who1 & 0.001\\
& \textbf{CO} & \textbf{0.194} & HE x inc3 & 0.000 & CO x who2 & 0.001\\
& \textbf{HE} & \textbf{0.057} & HE x inc4 & 0.000 & CO x who3 & 0.001\\
& ASC x who1 & 0.002 & log(HE) x inc1 & 0.000 & TT x pur1 & 0.000\\
& \multicolumn{2}{c|}{...}& \multicolumn{2}{c|}{...}& \multicolumn{2}{c}{...}\\
\end{tabular}
\end{center}
\label{table:fake_results_hard1}
\end{table}%

Let us now consider the large search space. Table~\ref{table:fake_results_hard1} shows the top-K variables with higher $\lambda$ value according to DCM-ARD for S1, S2 and S3. As the obtained results show, DCM-ARD is still able to recover the true specifications that were used to generate the data regardless of the significantly larger search space (602 variables considered, instead of 252 for Table~\ref{table:fake_results_easy1}). However, since the number of variables considered is substantially larger, the execution time of the proposed DSVI algorithm naturally increased from approximately 10 minutes to close to 1 hour on a standard 2.3 GHz dual-core laptop with 16 GB of RAM. 

\begin{table}[t]
\caption{Results of DCM-ARD for large search space (part 2).}
\begin{center}
\setlength\tabcolsep{5pt}
\begin{tabular}{c| lc | lc | lc}
& \multicolumn{2}{c|}{Train} & \multicolumn{2}{c|}{Swiss Metro} & \multicolumn{2}{c}{Car}\\
\hline
 Spec & Variable & $\lambda$ & Variable & $\lambda$ & Variable & $\lambda$\\
\hline
\multirow{6}{*}{S7} & \textbf{ASC} & \textbf{2.246} & \textbf{TT} & \textbf{0.574} & \textbf{TT} & \textbf{0.553}\\
& \textbf{box(TT) x ga} & \textbf{1.787} & \textbf{ASC} & \textbf{0.120} & \textbf{CO} & \textbf{0.019}\\
& \textbf{CO} & \textbf{0.360} & CO x pur1 & 0.000 & TT x lugg1 & 0.000\\
& \textbf{log(TT)} & \textbf{0.220} & CO x pur2 & 0.000 & TT x lugg2 & 0.000\\
& log(CO) x inc1 & 0.001 & CO x pur3 & 0.000 & seg(CO,4) & 0.000\\
& \multicolumn{2}{c|}{...}& \multicolumn{2}{c|}{...}& \multicolumn{2}{c}{...}\\
\hline
\multirow{9}{*}{S8} & \textbf{ASC x ga} & \textbf{7.448} & \textbf{ASC x ga} & \textbf{4.805} & \textbf{TT} & \textbf{0.828}\\
& \textbf{CO} & \textbf{2.840} & \textbf{CO} & \textbf{1.695} & \textbf{CO} & \textbf{0.018}\\
& \textbf{ASC} & \textbf{1.611} & \textbf{TT} & \textbf{0.559} & seg(CO,4) & 0.000\\
& \textbf{TT} & \textbf{1.120} & \textbf{ASC} & \textbf{0.336} & seg(CO,4) & 0.000\\
& \textbf{CO x who1} & \textbf{0.057} & \textbf{CO x who1} & \textbf{0.025} & seg(CO,4) & 0.000\\
& \textbf{CO x who2} & \textbf{0.057} & \textbf{CO x who2} & \textbf{0.025} & CO x pur1 & 0.000\\
& \textbf{CO x who3} & \textbf{0.057} & \textbf{CO x who3} & \textbf{0.025} & CO x pur2 & 0.000\\
& CO x inc1 & 0.001 & seg(TT,8) & 0.000 & CO x pur3 & 0.000\\
& \multicolumn{2}{c|}{...}& \multicolumn{2}{c|}{...}& \multicolumn{2}{c}{...}\\
\hline
\multirow{10}{*}{S9} & \textbf{ASC} & \textbf{2.255} & \textbf{TT} & \textbf{1.367} & \textbf{TT} & \textbf{1.118}\\
& \textbf{TT} & \textbf{1.197} & \textbf{CO x ga} & \textbf{0.501} & \textbf{CO} & \textbf{0.098}\\
& \textbf{CO x ga} & \textbf{0.805} & \textbf{TT x age1} & \textbf{0.134} & \textbf{CO x inc1} & \textbf{0.004}\\
& \textbf{CO} & \textbf{0.187} & \textbf{TT x age2} & \textbf{0.134} & \textbf{CO x inc2} & \textbf{0.004}\\
& ASC x who1 & 0.001 & \textbf{TT x age3} & \textbf{0.134} & \textbf{CO x inc3} & \textbf{0.004}\\
& ASC x who2 & 0.001 & \textbf{TT x age4} & \textbf{0.134} & \textbf{CO x inc4} & \textbf{0.004}\\
& ASC x who3 & 0.001 & \textbf{ASC} & \textbf{0.110} & CO x who1 & 0.001\\
& HE x age1 & 0.000 & \textbf{CO} & \textbf{0.015} & CO x who2 & 0.001\\
& HE x age2 & 0.000 & seg(TT,8) & 0.000 & CO x who3 & 0.001\\
& \multicolumn{2}{c|}{...}& \multicolumn{2}{c|}{...}& \multicolumn{2}{c}{...}\\
\end{tabular}
\end{center}
\label{table:fake_results_hard2}
\end{table}%

Lastly, Table~\ref{table:fake_results_hard2} shows the top-K variables deemed relevant by DCM-ARD for inclusion in the utility function specifications for S7, S8 and S9. By comparing these results with the true specifications from Table~\ref{table:fakespecs}, one can again observe that DCM-ARD is able to discover the true specifications almost exactly. The only differences are the fact that DCM-ARD selected ``log(TT)" instead of ``box(TT)" in the utility function of train in S7, and the fact that it missed the interaction between ``CO" and ``luggage" in the utility function of car in S8. While we could not find an obvious explanation for the latter, the former can be easily explained by an analysis of the results of the Box-Cox transform, which uses a maximum likelihood approach to fit the parameters of the transformation. In the particular case of train travel time, we could immediately observe that the transformed values produced by the Box-Cox transformation are almost perfectly correlated with to the ones produced by the log-transformation (correlation coefficient of 0.998), thus leading us to conclude that both lead to equivalent utility function specifications for the train alternative. 

As a further test of scalability and robustness of the proposed approach, we also considered an extremely large search space, which was obtained by expanding the large space space described above with variables that consist of Gaussian random noise, until a total of 1000 variables per alternative was reached (\mbox{i.e.}, a total of 3000 variables). Using the semi-artificial choice data corresponding to specification S2 we were able to verify that, despite the expected increased computational run time (approximately 5 hours), the proposed DCM-ARD was still able to perfectly recover the true specification of S2. 

So far we have only been considering the ability of DCM-ARD to infer the correct utility function specifications. However, one can also evaluate DCM-ARD in terms of its prediction accuracy on held-out data. Table~\ref{table:pred_acc} shows the prediction accuracy of DCM-ARD when trained only on 70\% of the dataset and tested on 30\% held-out data for the different semi-artificial specifications considered (S1-S9). By comparing these results with the accuracy of a standard DCM that considers all the variables from the search space as input (``DCM"), one can verify that thanks to the additional flexibility of the proposed hierarchical prior and the sparsity-inducing properties, DCM-ARD is able to generalize better to held-out data, thus resulting in significantly higher prediction accuracies. In fact, is most cases, DCM-ARD achieves almost as good prediction performance as a DCM estimated using the true specifications that were used to generate the semi-artificial choices (``DCM-TRUE"). On the other hand, a DCM fitted with maximum likelihood estimation with such a high number of input variables is very likely to severely overfit.

\begin{table}[t]
\caption{Prediction accuracy and log-likelihood on held-out data}
\begin{center}
\begin{tabular}{c | c | c c | c c | c c}
 & & \multicolumn{2}{c|}{DCM} & \multicolumn{2}{c|}{DCM-ARD} & \multicolumn{2}{c}{DCM-TRUE}\\
 \hline
Search Space & Spec & Acc. & LogLik & Acc. & LogLik & Acc. & LogLik\\
\hline
Moderate & S1 & 0.615 & -2733.9 & 0.628 & -2569.0 & 0.627 & -2567.4\\
Moderate & S2 & 0.627 & -2697.0 & 0.638 & -2498.2 & 0.636 & -2496.8\\
Moderate & S3 & 0.639 & -2662.5 & 0.645 & -2452.9 & 0.646 & -2450.4\\
Moderate & S4 & 0.627 & -2597.3 & 0.647 & -2454.9 & 0.648 & -2452.7\\
Moderate & S5 & 0.607 & -2788.8 & 0.623 & -2621.9 & 0.623 & -2619.2\\
Moderate & S6 & 0.624 & -2621.3 & 0.632 & -2530.3 & 0.633 & -2527.1\\
\hline
Large & S1 & 0.589 & -2798.2 & 0.628 & -2569.0 & 0.627 & -2567.4\\
Large & S2 & 0.602 & -2773.7 & 0.638 & -2498.2 & 0.636 & -2496.8\\
Large & S3 & 0.612 & -2924.0 & 0.645 & -2452.9 & 0.646 & -2450.4\\
Large & S7 & 0.603 & -2746.6 & 0.606 & -2675.5 & 0.617 & -2551.1\\
Large & S8 & 0.598 & -2858.0 & 0.642 & -2489.8 & 0.646 & -2421.7\\
Large & S9 & 0.614 & -2823.9 & 0.653 & -2466.7 & 0.660 & -2400.3\\
\end{tabular}
\end{center}
\label{table:pred_acc}
\end{table}%

\begin{table}[t!]
\caption{Results for real SM data}
\begin{center}
\setlength\tabcolsep{5pt}
\begin{tabular}{lc | lc | lc}
\multicolumn{2}{c|}{Train} & \multicolumn{2}{c|}{Swiss Metro} & \multicolumn{2}{c}{Car}\\
\hline
 Variable & $\lambda$ & Variable & $\lambda$ & Variable & $\lambda$\\
\hline
log(TT) x ga & 9.506 & log(CO) x ga & 5.570 & log(CO) & 4.479\\
ASC & 4.002 & log(CO) x pur1 & 2.251 & TT x ga & 1.378\\
log(CO) & 3.262 & log(CO) x pur2 & 2.251 & log(TT) x pur1 & 0.477\\
log(CO) x pur1 & 2.469 & log(CO) & 1.184 & log(TT) x pur2 & 0.477\\
log(CO) x pur2 & 2.469 & log(TT) & 0.506 & log(CO) x age1 & 0.213\\
log(CO) x ga & 1.235 & CO & 0.349 & log(CO) x age2 & 0.213\\
CO & 0.556 & ASC x age1 & 0.250 & log(CO) x age3 & 0.213\\
log(CO) x age1 & 0.269 & ASC x age2 & 0.250 & log(CO) x age4 & 0.213\\
log(CO) x age2 & 0.269 & ASC x age3 & 0.250 & CO x pur1 & 0.156\\
log(CO) x age3 & 0.269 & ASC x age4 & 0.250 & CO x pur2 & 0.156\\
log(CO) x age4 & 0.269 & CO x pur1 & 0.236 & TT x age1 & 0.107\\
CO x pur1 & 0.228 & CO x pur2 & 0.236 & TT x age2 & 0.107\\
CO x pur2 & 0.228 & ASC x ga & 0.146 & TT x age3 & 0.107\\
log(TT) & 0.175 & CO x ga & 0.099 & TT x age4 & 0.107\\
log(HE) & 0.075 & TT x age1 & 0.027 & CO & 0.037\\
CO x ga & 0.068 & TT x age2 & 0.027 & log(TT) & 0.000\\
CO x age1 & 0.034 & TT x age3 & 0.027 & log(TT) x ga & 0.000\\
CO x age2 & 0.034 & TT x age4 & 0.027 & log(CO) x ga & 0.000\\
CO x age3 & 0.034 & TT x pur1 & 0.005 & TT & 0.000\\
CO x age4 & 0.034 & TT x pur2 & 0.005 & TT x pur1 & 0.000\\
\multicolumn{2}{c|}{...}& \multicolumn{2}{c|}{...}& \multicolumn{2}{c}{...}\\
\end{tabular}
\end{center}
\label{table:results_ard_real}
\end{table}%

\subsection{Real choice data}

\begin{table}[t]
\caption{Utility function specifications for true SM data}
\begin{center}
\setlength\tabcolsep{4pt}
\begin{tabular}{c | l | l | l}
& \multicolumn{3}{c}{Specification}\\
\hline
S\# & Variables in $V_{\mbox{\scriptsize train}}$ & Variables in $V_{\mbox{\scriptsize sm}}$ & Attrib. in $V_{\mbox{\scriptsize car}}$\\
\hline
\hline
R1 & ASC, TT, CO & ASC, TT, CO & TT, CO \\
\hline
\multirow{2}{*}{R2} & ASC, \textbf{log(TT)}, & ASC, \textbf{log(TT)},  & TT, \textbf{log(CO)} \\
 & \textbf{log(TT) x ga}, \textbf{log(CO)} & \textbf{log(CO)} \\
\hline
\multirow{2}{*}{R3} & ASC, log(TT), log(TT) x ga, & ASC, log(TT),  log(CO), & TT, log(CO) \\
 & log(CO), \textbf{log(CO) x pur} & \textbf{log(CO) x ga} \\
\hline
\multirow{3}{*}{R4} & ASC, log(TT), log(TT) x ga, & ASC, log(TT),  & TT, \textbf{TT x ga}, \\
 & log(CO), \textbf{log(CO) x ga}, & log(CO), log(CO) x ga, & log(CO) \\
 & log(CO) x pur & \textbf{log(CO) x pur} \\
\hline
\multirow{4}{*}{R5} & ASC, log(TT), log(TT) x ga, & ASC, \textbf{ASC x age}, & TT, TT x ga, \\
 & log(CO), log(CO) x ga, & log(TT), log(CO), & \textbf{TT x pur}, \\
 & log(CO) x pur, & log(CO) x ga, & log(CO) \\
 & \textbf{log(CO) x age} & log(CO) x pur \\
\hline
\multirow{4}{*}{R6} & ASC, log(TT), log(TT) x ga, & ASC, ASC x age, & TT, TT x ga, \\
 & log(CO), log(CO) x ga, & log(TT), log(CO), & TT x pur, \\
 & log(CO) x pur, & log(CO) x ga, & log(CO) \\
 & log(CO) x age, \textbf{log(HE)} & log(CO) x pur \\
\hline
\multirow{4}{*}{R7} & ASC, log(TT), log(TT) x ga, & ASC, \textbf{ASC x ga}, & TT, TT x ga, \\
 & log(CO), log(CO) x ga, & ASC x age, log(TT), & TT x pur, log(CO) \\
 & log(CO) x pur, & log(CO), log(CO) x ga, & \textbf{log(CO) x age}, \\
 & log(CO) x age, log(HE) & log(CO) x pur & \textbf{log(CO) x pur} \\
\end{tabular}
\end{center}
\label{table:real_specs}
\end{table}%

\begin{table}[t!]
\caption{Results for true SM data}
\begin{center}
\setlength\tabcolsep{5.5pt}
\begin{tabular}{l | c c c c c c c}
& \multicolumn{7}{c}{Specification}\\
& R1 & R2 & R3 & R4 & R5 & R6 & R7\\
\hline
Log-like full & -8,625 & -8,368 & -8,064 & -7,836 & -7,679 & -7,645 & \textbf{-7,617}\\
AIC & 17,267 & 16,755 & 16,152 & 15,704 & 15,410 & 15,345 & \textbf{15,301}\\
BIC & 17,326 & 16,821 & 16,239 & 15,820 & 15,599 & \textbf{15,542} & 15,549\\
Pseudo-$R^2$ & 0.221 & 0.244 & 0.272 & 0.292 & 0.306 & 0.309 & \textbf{0.312}\\
Pseudo-$\bar{R}^2$ & 0.220 & 0.243 & 0.271 & 0.291 & 0.304 & 0.307& \textbf{0.309}\\
\hline
Log-lik train & -6,032 & -5,822 & -5,619 & -5,429 & -5,297 & -5,271 & \textbf{-5,247}\\
Log-lik test & -2,603 & -2,558 & -2,457 & -2,437 & -2,428 & \textbf{-2,421} & 2,430\\
Train acc. & 0.616 & 0.636 & 0.661 & 0.676 & 0.689 & 0.690 & \textbf{0.692}\\
Test acc. & 0.615 & 0.638 & 0.662 & 0.670 & 0.675 & 0.677 & \textbf{0.679}\\
\end{tabular}
\end{center}
\label{table:results_dcm_true}
\end{table}%

We will now consider the application of DCM-ARD to perform automatic utility function specification on the real choice data from the Swissmetro dataset. Table~\ref{table:results_ard_real} shows the top-20 variables selected by DCM-ARD for inclusion in the utility functions using the moderate-sized search space. Since in this case the correct specification is unknown, we instead evaluate the quality of the DCM models that the specifications inferred by DCM-ARD produce. With this purpose, we developed a series of specifications of increasing complexity based on the results of Table~\ref{table:results_ard_real}. We begin by considering a rather simplistic specification based only on travel time and cost (R1). We then start adding variables to it according to the results of DCM-ARD in descending order of importance according to the learned values of $\lambda$. The complete set of specifications considered is show in Table~\ref{table:real_specs}. Kindly note that the last specification (R7), already includes almost all the variables in the top-20 ranking shown in Table~\ref{table:results_ard_real}, and that other additional variables were assigned a $\lambda$ value of zero (or very close to zero), thus being deemed irrelevant by DCM-ARD. Also, since including both a variable and its log-transform could compromise the interpretability of the DCM models, we decided to include only the version with the higher value of $\lambda$ in the cases where DCM-ARD selected both variants\footnote{We note that, according to our empirical evidence, including both variants does tend to lead to models that fit better the data, including the held-out data.}. Also, due to the fact that the purpose variable has 9 categories, with some of them having only a couple of observations, we further grouped the trip purposes into: commuting, shopping and leisure.

\begin{table}[t]
\vspace{0.5cm}
\caption{Results for true SM data vs. baseline from state of the art}
\begin{center}
\setlength\tabcolsep{5.5pt}
\begin{tabular}{l | c c c c }
& \multicolumn{4}{c}{Specification}\\
& \cite{bierlaire2001acceptance} & PyLogit Example & R6 & R7\\
\hline
Log-lik full & -8,483 & -8,061 & -7,645 & \textbf{-7,617}\\
AIC & 16,984 & 16,150 & 15,345 & \textbf{15,301}\\
BIC & 17,050 & 16,252 & \textbf{15,542} & 15,549\\
Pseudo-$R^2$ & 0.234 & 0.272 & 0.309 & \textbf{0.312}\\
Pseudo-$\bar{R}^2$ & 0.233 & 0.271 & 0.307& \textbf{0.309}\\
\hline
Log-lik train & -5,960 & -5,633 & -5,271 & \textbf{-5,247}\\
Log-lik test & -2,535 & -2,450 & \textbf{-2,421} & 2,430\\
Train acc. & 0.646 & 0.667 & 0.690 & \textbf{0.692}\\
Test acc. & 0.644 & 0.650 & 0.677 & \textbf{0.679}\\
\end{tabular}
\end{center}
\label{table:results_dcm_true_literature}
\vspace{-.2cm}
\end{table}%

Based on the specifications that were generated according to the results of DCM-ARD (Table~\ref{table:results_ard_real}), we then fitted standard DCM models using the PyLogit package \citep{brathwaite2018asymmetric} in Python. Table~\ref{table:results_dcm_true} shows the results obtained for the different specifications considered. As expected, one can verify that, as we increase the complexity of the specification according to the results of DCM-ARD, the fit of the DCM model improves in terms of log-likelihood. However, the quality of the DCM model also improves in terms of AIC, BIC and pseudo-$\bar{R}^2$. In order to further assess the quality of the DCM-ARD specifications in terms of generalization ability to held-out data, we also performed a random 70/30\% train/test split of the dataset, and computed the likelihood and accuracies in both sets. As the results in Table~\ref{table:results_dcm_true} evidence, as we move towards the full specification inferred by DCM-ARD, the accuracy and held-out data likelihood of the DCM model also improves. Interestingly, it can observed that only when we include essentially all the variables deemed relevant by DCM-ARD we start noticing some signs of overfitting in the standard DCM model: BIC and testset likelihood do not improve when going from specification R6 to R7. However, indicators such as AIC and pseudo-$\bar{R}^2$ still improve. Furthermore, it should be noted that the variables included from R6 to R7, already consist of variables for which DCM-ARD  assigned a relatively low relevance (\mbox{i.e.} low value of $\lambda$ when compared to the others).

\begin{table}[t!]
\caption{Results for true SM data, spec 6}
\begin{center}
\setlength\tabcolsep{5.pt}
\begin{tabular}{l | c c c c c c}
 & Coef & StdErr & $z$ & $p>|z|$ & [0.025 & 0.975] \\
\hline
ASC (Train) & 3.036 & 0.196 & 15.478 & 0.000 & 2.652 & 3.421 \\
ASC (SM) & 0.900 & 0.134 & 6.725 & 0.000 & 0.638 & 1.163 \\
ASC x age1 (SM) & 0.575 & 0.156 & 3.699 & 0.000 & 0.271 & 0.880 \\
ASC x age2 (SM) & 0.784 & 0.103 & 7.585 & 0.000 & 0.582 & 0.987 \\
ASC x age3 (SM) & 0.704 & 0.102 & 6.909 & 0.000 & 0.505 & 0.904 \\
ASC x age4 (SM) & 0.479 & 0.107 & 4.478 & 0.000 & 0.270 & 0.689 \\
log(TT) (Train) & -0.964 & 0.261 & -3.697 & 0.000 & -1.477 & -0.453 \\
log(TT) (SM) & -2.570 & 0.110 & -23.465 & 0.000 & -2.785 & -2.355 \\
TT (Car) & -0.865 & 0.218 & -3.974 & 0.000 & -1.293 & -0.439 \\
log(TT) x ga (Train) & -2.995 & 0.275 & -10.880 & 0.000 & -3.535 & -2.455 \\
TT x ga (Car) & -0.176 & 0.210 & -0.841 & 0.400 & -0.589 & 0.235 \\
TT x pur1 (Car) & 0.273 & 0.064 & 4.285 & 0.000 & 0.148 & 0.398 \\
TT x pur2 (Car) & 0.289 & 0.088 & 3.289 & 0.001 & 0.117 & 0.463 \\
log(CO) (Train) & -2.637 & 0.318 & -8.297 & 0.000 & -3.261 & -2.015 \\
log(CO) (SM) & -1.984 & 0.247 & -8.023 & 0.000 & -2.470 & -1.500 \\
log(CO) (Car) & -1.875 & 0.175 & -10.714 & 0.000 & -2.218 & -1.532 \\
log(CO) x ga (Train) & -1.997 & 0.195 & -10.248 & 0.000 & -2.379 & -1.615 \\
log(CO) x ga (SM) & -2.249 & 0.132 & -17.024 & 0.000 & -2.509 & -1.991 \\
CO x age1 (Train) & -0.317 & 0.090 & -3.539 & 0.000 & -0.493 & -0.141 \\
CO x age2 (Train) & -0.578 & 0.079 & -7.336 & 0.000 & -0.733 & -0.424 \\
CO x age3 (Train) & -0.647 & 0.080 & -8.134 & 0.000 & -0.804 & -0.492 \\
CO x age4 (Train) & -0.525 & 0.083 & -6.301 & 0.000 & -0.690 & -0.362 \\
log(CO) x pur1 (Train) & 2.521 & 0.294 & 8.574 & 0.000 & 1.945 & 3.098 \\
log(CO) x pur1 (SM) & 1.963 & 0.231 & 8.510 & 0.000 & 1.511 & 2.415 \\
log(CO) x pur2 (Train) & 3.282 & 0.308 & 10.641 & 0.000 & 2.678 & 3.887 \\
log(CO) x pur2 (SM) & 2.589 & 0.244 & 10.606 & 0.000 & 2.111 & 3.068 \\
HE, (Train) & -0.948 & 0.118 & -8.059 & 0.000 & -1.179 & -0.718 \\
\end{tabular}
\end{center}
\label{table:results_pylogit_real}
\vspace{-.2cm}
\end{table}%

Comparing the results of specifications R6 and R7 with other proposed DCM specifications from the literature for the same dataset (Table~\ref{table:results_dcm_true_literature}), it is possible to obtain a better perspective of how good the  specifications inferred by DCM-ARD are. For example, the DCM specification proposed in PyLogit for the Swissmetro dataset includes variables such as travel time, cost, headway, seat configuration, luggage and first class. However, it only achieves a loglikelihood of $-8,061$, a BIC of $16,252$ and a pseudo-$\bar{R}^2$ of $0.271$. Similarly, the original specification proposed by \cite{bierlaire2001acceptance} achieves a loglikelihood of just $-8,483$, a BIC of $17,050$ and a pseudo-$\bar{R}^2$ of $0.233$. Moreover, if we consider generalization to held-out data, Table~\ref{table:results_dcm_true_literature} also demonstrates that the both R6 and R7 obtain better results than both baseline approaches, thereby highlighting how DCM-ARD can be easily used to enable the automatic search of utility function specifications.

Lastly, Table~\ref{table:results_pylogit_real} shows the estimated coefficients by a DCM with the specification R6 using PyLogit, and their corresponding p-values and other statistics. The full set of results for the other specifications were omitted for brevity but are available at \url{http://fprodrigues.com/DCM-ARD/}, together with the source code. As the results in Table~\ref{table:results_pylogit_real} demonstrate, the specification learned by DCM-ARD leads to a stable DCM in which the coefficients for all variables except ``TT x ga (Car)'', have p-values smaller than $0.001$. It should however be noted that, in two cases, the parameter estimates are not entirely behaviourally realistic: for both Train and SM alternatives, the sum of the parameter related to ``log(CO) x pur2'' and the corresponding baseline (``log(CO) (Train)'' and ``log(CO) (SM)'') is positive, implying that all else being equal, increasing the travel cost of shopping trips improves their attractiveness. Such result is obviously wrong; it indicates that the involved parameters are erroneously capturing or omitting some effects, most probably because the travel cost of the two affected modes is interacted with ``ga'' and ``pur'', but not with both simultaneously. However, since such interactions were not considered in the search-space, DCM-ARD is unable to identify them as relevant.
Thus, this is a great example that highlights an important limitation of DCM-ARD: its results are dependent of the search-space considered, and it has no knowledge of behavioural theories. However, we reiterate that its purpose is to assist modellers on specifying utility functions according to data-driven knowledge, rather then serving as a replacement for expert modellers and domain knowledge.

\section{Conclusion}
\label{sec:conclusion}

This paper proposed a Bayesian framework for performing automatically utility function specification in discrete choice models based on the idea of automatic relevance determination (ARD). An efficient doubly stochastic variational inference algorithm was derived in order to perform approximate Bayesian inference in the proposed DCM-ARD model. As our empirical results using both semi-artificial and real choice data showed, the proposed approach is able to automatically discover good utility function specifications in a pure data-driven manner, even in situations when the number of possible variables considered for inclusion in the utility functions is very large. The practical advantages and overall feasibility of the proposed approach were demonstrated through an application to the popular Swissmetro dataset \citep{bierlaire2001acceptance}, where DCM-ARD was shown to be capable of generating specifications that outperform others from the state of the art according to multiple criteria. 

Despite the importance of the standard formulation of the multinomial logit in discrete choice theory, it only corresponds to a subset of the models that are used in practice, with modelling approaches like mixed logits and latent class choice models providing important ways of capturing the heterogeneity in preferences among the decision makers. Therefore, our future work focuses on extending the proposed DCM-ARD formulation for this type models, and on dealing with the challenges associated with performing approximate Bayesian inference in those settings in a scalable manner. 

\section*{References}

\bibliography{dcm-ard}

\end{document}